\documentclass[sigconf, screen]{acmart}

\AtBeginDocument{%
  }

\setcopyright{none}
\copyrightyear{2026}
\acmYear{2026}
\acmDOI{XXXXXXX.XXXXXXX}

\acmConference[arXiv]{arXiv}{2026}{}
\acmISBN{978-1-4503-XXXX-X/2026/07}
\renewcommand\footnotetextcopyrightpermission[1]{}
\usepackage{dsfont}
\usepackage[mathscr]{euscript}

\newcommand{\ra}[1]{\renewcommand{\arraystretch}{#1}} 
\usepackage[table]{xcolor}
\usepackage{color, colortbl}
\definecolor{Gray}{gray}{0.9}
\definecolor{light-gray}{gray}{0.95}

\usepackage{pifont}
\newcommand{\cmark}{\ding{51}}%
\newcommand{\xmark}{\ding{55}}%

\usepackage{multirow} 
\usepackage{subcaption} 
\usepackage{graphicx} 

\usepackage{comment}
\usepackage{enumitem}

\newcommand{\eg}{\emph{e.g.}}

\begin{document}

\hypersetup{
  pagebackref=true,
  breaklinks=true,
  colorlinks=true,
  citecolor=black,
  linkcolor=black,
  urlcolor=teal
}

\title{SARAH: Spatially Aware Real-time Agentic Humans}

\author{Evonne Ng\quad Siwei Zhang\quad Zhang Chen\quad Michael Zollhoefer\quad Alexander Richard}

\affiliation{%
  \vspace{0.35em}
  \institution{Meta Reality Labs}
  \city{Redmond}
  \state{WA}
  \country{USA}
}

\renewcommand{\shortauthors}{Ng et al.}
\newcommand{\mname}[1]{Name}

\begin{abstract}
    As embodied agents become central to VR, telepresence, and digital human applications, their motion must go beyond speech-aligned gestures: agents should turn toward users, respond to their movement, and maintain natural gaze. Current methods lack this spatial awareness.
    We close this gap with the first real-time, fully causal method for spatially-aware conversational motion, deployable on a streaming VR headset.
    Given a user's position and dyadic audio, our approach produces full-body motion that aligns gestures with speech while orienting the agent according to the user.
    Our architecture combines a causal transformer-based VAE with interleaved latent tokens for streaming inference and a flow matching model conditioned on user trajectory and audio.
    To support varying gaze preferences, we introduce a gaze scoring mechanism with classifier-free guidance to decouple learning from control: the model captures natural spatial alignment from data, while users can adjust eye contact intensity at inference time.
    On the Embody 3D dataset, our method achieves state-of-the-art motion quality at over 300 FPS—3$\times$ faster than non-causal baselines—while capturing the subtle spatial dynamics of natural conversation.
    We validate our approach on a live VR system, bringing spatially-aware conversational agents to real-time deployment. See our \href{https://evonneng.github.io/sarah/}{project page} for details.
\end{abstract}

\begin{teaserfigure}
  \includegraphics[width=\textwidth]{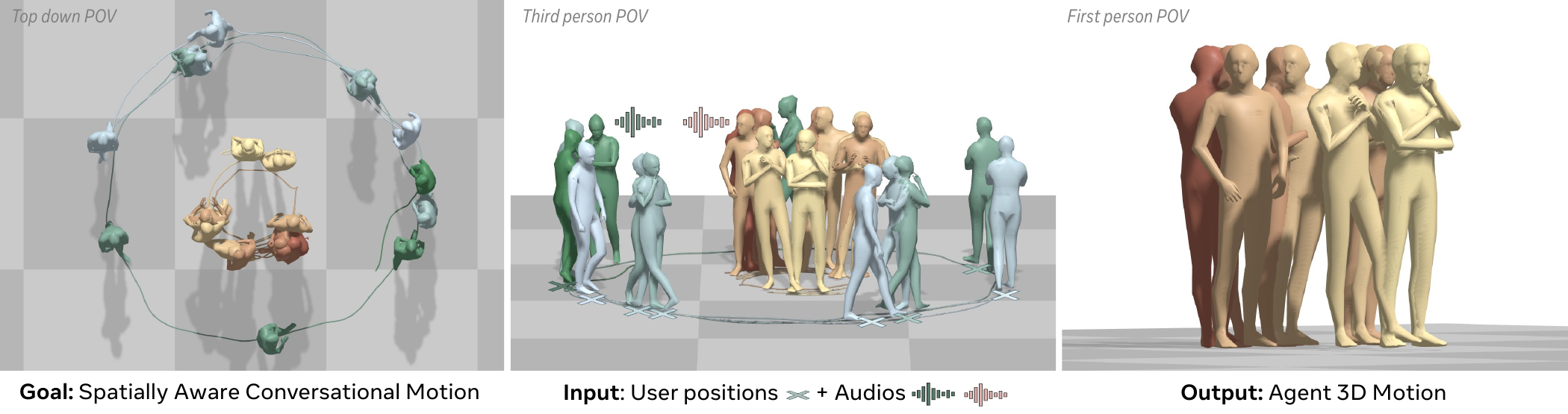}
  \caption{
  Our method generates full-body 3D motion for a virtual agent that is spatially aware of the user while engaging in a conversation.
  Given the user's floor-projected head trajectory and dyadic audio, we generate the agent's complete 3D motion.
  Trajectory colors indicate time: blue $\rightarrow$ green (user) and yellow $\rightarrow$ red (agent). See \href{https://evonneng.github.io/sarah/}{project page} for results.}
  \label{fig:teaser}
  \vspace{1em}
\end{teaserfigure}

\maketitle

\section{Introduction}
\label{sec:intro}

Embodied conversational agents are becoming central to immersive applications—from virtual reality companions and telepresence avatars to social robots and digital humans.
For these agents to feel truly present, speech alone is not enough.
Consider interacting with an agent that only stares forward as you walk around it, or an agent that wanders off as you are mid-sentence. 
Such behavior immediately breaks the illusion of presence.
Humans naturally turn toward their conversational partners, shift posture as they move, and modulate gaze to signal engagement.
Moreover, comfort in levels of eye contact vary widely—shaped by personal preference, social context, and cultural norms.
For virtual agents to replicate this behavior and appear humanlike, their motion must be both \emph{spatially aware} and \emph{controllable}—orienting toward the user while adapting gaze to individual preferences.
Current methods, however, focus on conversational contexts in isolation, producing agents that lack situated reasoning.

We present a method for generating full-body motion for a virtual agent that responds to both the conversation and the user's spatial movement—all in real-time.
Achieving such motion requires satisfying four criteria simultaneously.
First, it must be \emph{conversationally appropriate}—gestures should align naturally with speech.
Second, it must be \emph{spatially aware}—the agent should orient toward and react to the user's movement.
Third, it must be \emph{controllable}—gaze engagement should be adjustable to suit different contexts and preferences.
Fourth, it must be \emph{real-time}—generation must be causal and streaming, with no access to future information.
Achieving all four remains an open challenge: state-of-the-art methods either ignore spatial context, require non-causal access to future frames, or run far below real-time speeds.
We present the first method to close this gap.

Existing gesture generation methods are predominantly monadic: they synthesize motion for a single speaker conditioned on audio or text, with no awareness of an interlocutor~\cite{nyatsanga2023comprehensive, yi2023generating, alexanderson2023listen}.
The few dyadic methods that exist typically assume stationary, forward-facing participants—mimicking video calls rather than dynamic, in-person interactions~\cite{ng2024audio, ng2022learning}.
Moreover, popular state-of-the-art generative models are often too slow for real-time deployment~\cite{ng2024audio, ng2022learning} or require non-causal access to future frames~\cite{alexanderson2023listen}, precluding streaming inference.
Compounding this, existing dyadic datasets lack the spatial dynamics needed to learn reactive behavior. 
As a result, generated agents remain stationary and rigidly face one another—lacking the fluid spatial dynamics of real conversation.

Our key insight is to decouple learning from control: we learn the natural distribution of spatial alignment from data, capturing gaze behaviors from sustained eye contact to deliberate aversion, then apply a lightweight guidance mechanism at inference to calibrate orientation based on user preference.
This separation allows the model to generate motion that is both naturalistic (drawn from the learned distribution) and controllable (steered toward a desired gaze intensity).
To achieve this, we propose a real-time, causal architecture built on two core components.
First, a causal transformer-based VAE compresses motion into a temporally-strided latent sequence, with interleaved latent tokens enabling streaming inference without sacrificing temporal coherence.
Second, a flow matching model generates motion in this latent space, conditioned on the user's trajectory and both speakers' audio.
For fine-grained control, we introduce a gaze guidance mechanism based on classifier-free guidance, allowing users to modulate eye contact intensity at inference.
Underpinning these components is a fully Euclidean motion representation that improves training stability and enables precise end-effector control.

We evaluate on the Embody 3D dataset~\cite{mclean2025embody3d}, the first to capture realistic proxemics in dynamic spatial interactions.
Our method achieves state-of-the-art motion quality while running at over 300 FPS, outperforming non-causal baselines (MDM, A2P) that are 3$\times$ slower.
Notably, we match the gaze alignment of non-causal methods without access to future user positions, demonstrating that reactive spatial behavior can be learned causally.
The generated motion is also controllable: users can modulate eye contact intensity at inference to suit their preferences.
We deploy on a real-time avatar system, confirming viability for production.

In summary, we present the first real-time system for spatially-aware conversational motion, enabling virtual agents to participate in dynamic interactions.
Our approach combines a causal transformer-based VAE with interleaved latent tokens for streaming inference, a Euclidean surface-point representation for stable training and precise end-effector control, and a classifier-free gaze guidance mechanism for user-adjustable eye contact.
We achieve state-of-the-art performance on the Embody 3D dataset~\cite{mclean2025embody3d} and successfully deploy our method on a real-time avatar system.
\begin{figure*}[t]
    \centering
    \includegraphics[width=\textwidth]{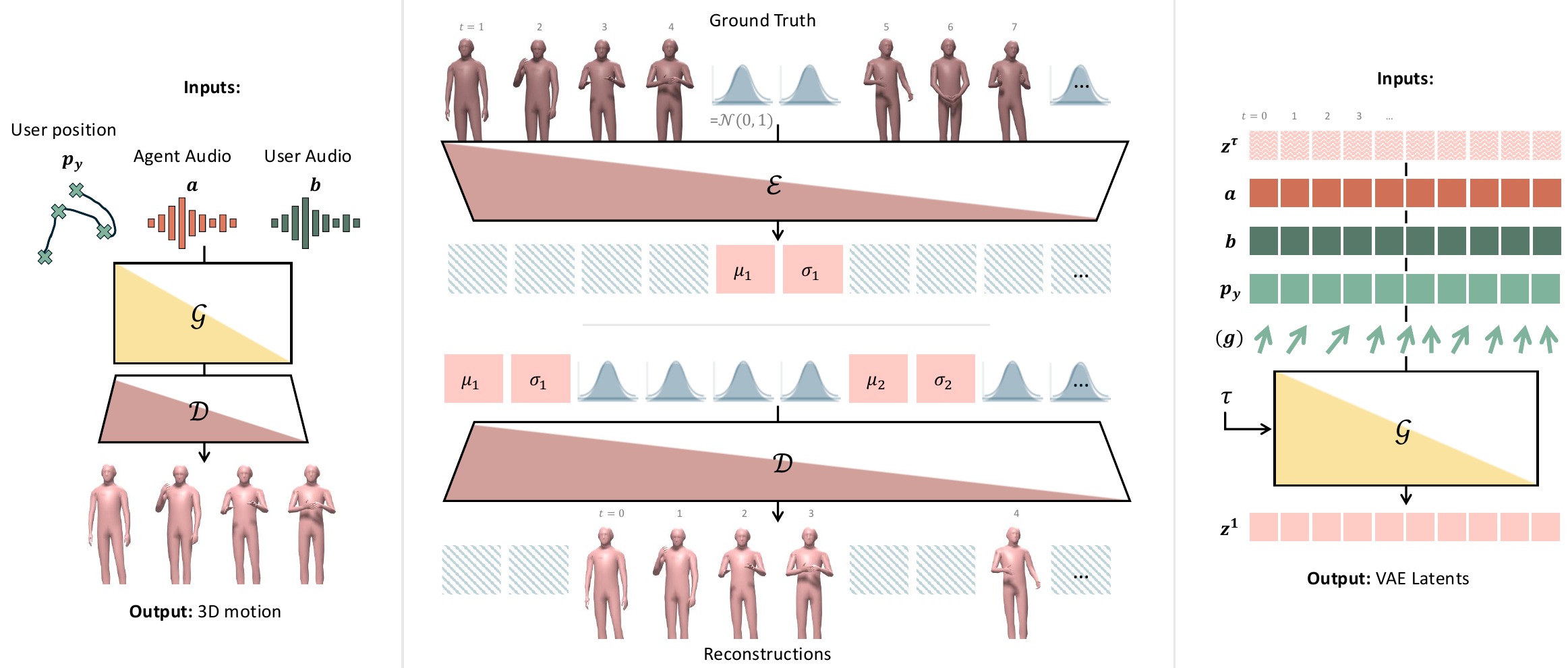}
    \caption{Given the user's 3D position and dyadic conversational audio, our model generates 3D motion that is conversationally and spatially aware (left). 
    We use a fully causal transformer-based VAE with interleaved latent tokens at a fixed temporal stride; both encoder and decoder employ causal attention, where each $\mu/\sigma$ token attends only to preceding frames and earlier latents (center). 
    These latents are passed to a transformer-based flow matching model that also uses causal masking and optionally accepts a gaze score for controlling the agent's eye contact (right). 
    Our lightweight architecture enables real-time, autoregressive streaming without distillation.}
    \label{fig:method}
\end{figure*}

\section{Related work}
\label{sec:relwork}
\subsection{Gestural motion generation.}
Most prior work on gestural motion generation has focused on single-person, co-speech gesture synthesis~\cite{nyatsanga2023comprehensive}, generating gestures that align with speaker audio.
Early methods employed recurrent neural networks~\cite{ghorbani2023zeroeggs} and feed-forward architectures~\cite{kucherenko2020gesticulator, ginosar2019learning}. 
More recent approaches use autoregressive transformers to produce vector-quantized motion tokens~\cite{yi2023generating} that decode into continuous motion. 
Conditional diffusion models have also become prominent~\cite{alexanderson2023listen, ao2023gesturediffuclip, yu2023talking, Zhi_2023_ICCV, liu2024tango}. 
Beyond audio, recent work has investigated text- and semantics-based conditioning for stylized gesture generation~\cite{cheng2024siggesture, zhang2024semantic}.
However, all of these works notably focus only on speakers in monadic settings. 


\subsection{Proxemics in interpersonal communication}
Oculesics (eye gaze and contact~\cite{kendon1967some}) and proxemics (interpersonal distance~\cite{argyle1965eye}) play crucial roles in regulating turn-taking, signaling attention, and communicating intent.
These signals have been used as priors for predicting social formations~\cite{alahi2016social}, trajectory forecasting~\cite{xie2024pedestrian, yang2024ia}, egocentric pose estimation~\cite{ng2020you2me, zhang2022egobody}, social behavior analysis~\cite{ref53}, and activity recognition~\cite{ref52, ref61, huang2014action}. Unlike methods that use oculesic and proxemic information as priors, we directly predict these signals.


Fine-grained gaze and head motion modeling has been studied for dyadic conversational motion~\cite{ng2022learning, ng2024audio, ahuja2019react, lee2019talking}. 
However, many focus on forward-facing video calls where proxemic information is lost~\cite{ng2022learning, ng2024audio}, or use datasets where dyadic pairs remain stationary~\cite{lee2019talking, ahuja2019react}. 
Due to scarce datasets capturing global proxemics, recent approaches leverage LLMs to reason about proxemic cues via language. 
For example, \cite{10.1145/3757377.3763879} uses an LLM for high-level gaze, proxemics, and pose guidance in dyadic interactions, while \cite{subramanian2024pose} employs an LLM to refine poses of closely interacting individuals. 
In contrast, we adopt a supervised approach to directly learn fine-grained proxemic information. 
Closely related, \cite{joo2019towards} addresses gaze and turn-taking prediction but decomposes the problem into sub-tasks without full-body locomotion. 
This is the first work to explicitly model fine-grained proxemics in dynamic, interactive dyadic conversations.


\subsection{Realtime causal generative modeling.}

Recent advances in generative motion synthesis have focused on acausal methods, \eg vanilla diffusion~\cite{tevet2022human, alexanderson2023listen, zhong2024smoodi}, which require both past and future context and are unsuitable for real-time applications.
To address this, some approaches combine vector-quantization (VQ) with causal transformers for fast, autoregressive generation~\cite{jiang2023motiongpt, guo2024momask, liu2024emage}.

More recently, diffusion models have been adapted for causal generation via conditioning on past frames~\cite{zhao2024dartcontrol, chen2024taming} or diffusion forcing~\cite{chen2024diffusion}. However, these still require multiple evaluation steps, making them slower than real-time. The video diffusion community has adopted distillation to compress multi-step models into single-step models for real-time streaming~\cite{lin2025diffusion, kodaira2025streamdit}. Motivated by these advances, we introduce an autoregressive, single-step flow-based model for real-time motion streaming.


\section{Real-time, Auto-regressive Motion Synthesis}
\label{sec:method}

Given a user and AI agent in conversation, our goal is to generate the agent's motion conditioned on both individuals' audio and the user's motion. 
Let $\mathbf{x} \in \mathbb{R}^{T \times D_x}$ and $\mathbf{y} \in \mathbb{R}^{T \times D_x}$ denote the motion sequences of the agent and user respectively, where $T$ is the sequence length and $D_x$ is the motion dimension.
In headset-based systems, full body pose is often unavailable while head position is always accessible.
We therefore condition only on the user's floor projected head position $\mathbf{p}_y \in \mathbb{R}^{T \times 2}$, computed as the midpoint between the left and right eyes and projected to the ground.
Let $\mathbf{a}, \mathbf{b} \in \mathbb{R}^{T \times D_a}$ denote the audio features of agent and user, where $D_a$ is the audio dimension.
We model the generation as:
\begin{equation}
    \mathbf{x} = \mathcal{G}(\mathbf{p}_y, \mathbf{a}, \mathbf{b}),
\end{equation}
where $\mathcal{G}$ is our generative model. For audio conditioning, we extract HuBERT features~\cite{hsu2021hubert} from each audio stream to obtain $\mathbf{a}$ and $\mathbf{b}$.

\begin{figure}[t]
\centering
\includegraphics[width=\columnwidth]{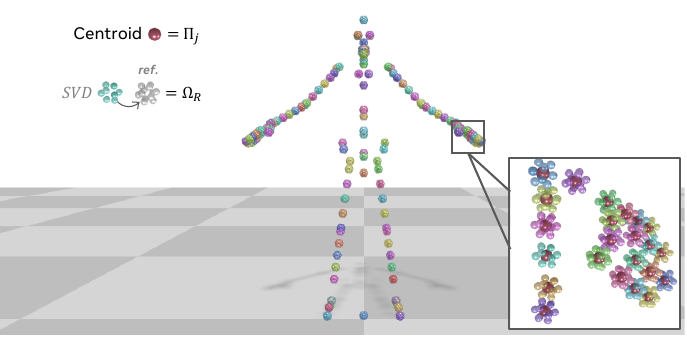}
\caption{We represent each joint $j$ as a 3D icosahedron. 
The centroid of the vertices yields the global position $\boldsymbol{\Pi}_j$, and we recover the global orientation $\boldsymbol{\Omega}_j$ via SVD against a reference icosahedron.}
\label{fig:representation}
\end{figure}

\subsection{Motion Representation}
\label{sec:motion-representation}
Traditionally, human motion is represented by local joint rotations $\boldsymbol{\theta}$ with root transforms $(R, \mathbf{t})$. 
Many methods predict $\boldsymbol{\theta}$ and $(R, \mathbf{t})$ directly, using forward kinematics and linear blend skinning to obtain meshes $\mathbf{M} \in \mathbb{R}^{T \times V \times 3}$.
We find that a fully Euclidean representation leads to faster convergence and more stable training.

To avoid error propagation from local rotations, we encode each joint $j$ as a 3D icosahedron: the centroid of its 12 vertices yields world-space position $\boldsymbol{\Pi}_j$, while SVD against a reference icosahedron recovers orientation $\boldsymbol{\Omega}_j$ (Fig.~\ref{fig:representation}).
Each pose is thus represented as $x_t \in \mathbb{R}^{J \times 12 \times 3}$, where $J$ is the number of joints.
We additionally include mesh $M_t$ as a shell around the joints to capture surface geometry.
To prevent unbounded drift, we normalize rotation and translation with respect to the first frame, aligning the agent at the origin facing the $z$-axis at $t{=}1$. 
As shown in Tab.~\ref{tab:baselines}, this representation leads to improved performance over traditional joint-angle parameterizations.

\subsection{Causal Transformer-based VAE}
\label{sec:causal-vae}
We propose a causal VAE architecture to support streaming inference. 
Unlike typical transformer VAEs that place global latent tokens at sequence start (enabling bidirectional attention), we interleave latent tokens at a fixed temporal stride $s$.

Concretely, the encoder $\mathcal{E}$ receives input ordered as:
\begin{equation}
(\mathbf{x}_{1:s},\, \mu_1,\, \sigma_1,\, \mathbf{x}_{s+1:2s},\, \mu_2,\, \sigma_2,\, \ldots),
\end{equation}
where $\mu_k, \sigma_k \in \mathbb{R}^{D_z}$ are the mean and variance tokens for block $k$, and $D_z$ is the latent dimension.
We apply causal self-attention: each frame attends only to past frames, and each $\mu_k/\sigma_k$ token attends to preceding frames and earlier latent tokens.
The decoder $\mathcal{D}$ mirrors this pattern.
See Fig.~\ref{fig:method} for an overview.

We optimize the VAE with reconstruction and KL losses:
\begin{equation}
\label{eq:vae}
\mathcal{L}_\text{VAE} = \| \mathbf{x} - \hat{\mathbf{x}} \|_2^2 + \beta \sum_{k=1}^{K} \mathrm{KL}\big(q_\phi(z_k \mid \mathbf{x}_{1:ks}) \,\|\, \mathcal{N}(\mathbf{0}, \mathbf{I})\big),
\end{equation}
where $q_\phi(z_k \mid \mathbf{x}_{1:ks}) = \mathcal{N}(\mu_k, \sigma_k^2)$ is the approximate posterior, $\beta$ is the KL weight, $K = T/s$ is the number of blocks, $\hat{\mathbf{x}}$ is the reconstruction, and $z_k \in \mathbb{R}^{D_z}$ is the sampled latent for block $k$.
After training, we use the encoder to obtain the latent sequence $\mathbf{z} = (z_1, \ldots, z_K) \in \mathbb{R}^{K \times D_z}$.

\subsection{Motion Generator}
\label{sec:motion-gen}
We adopt a transformer-based flow matching model for real-time, causal motion generation.
Flow matching transports samples from noise $\boldsymbol{\epsilon} \sim \mathcal{N}(\mathbf{0}, \mathbf{I})$ to data by predicting a velocity field $\mathbf{v}_\theta(\mathbf{z}^\tau, \tau, \mathbf{c})$, where $\tau \in [0, 1]$ is flow time, $\mathbf{z}^\tau$ is the interpolated latent, and $\mathbf{c}$ denotes conditioning.

We condition on the user's head position $\mathbf{p}_y$ and both audio streams $\mathbf{a}, \mathbf{b}$, predicting the agent's latent $\mathbf{z} \in \mathbb{R}^{K \times D_z}$.
At flow time $\tau$, we form:
\begin{equation}
\mathbf{z}^\tau = \tau \mathbf{z} + (1-\tau) \boldsymbol{\epsilon}, \quad \boldsymbol{\epsilon} \sim \mathcal{N}(\mathbf{0}, \mathbf{I}).
\label{eq:z_tau}
\end{equation}
We concatenate $\mathbf{z}^\tau$ with conditioning $\mathbf{c} = [\mathbf{p}_y; \mathbf{a}; \mathbf{b}]$ along the channel dimension, applying modality-specific positional encodings.
During training, we enforce classifier free guidance dropping each modality independently with a 5 percent probability.
The flow timestep $\tau$ is injected via adaptive layer normalization~\cite{peebles2023scalable}.
Using $x_1$-prediction, we train:
\begin{equation}
\label{eq:flow_loss}
\mathcal{L}_\text{flow} = \mathbb{E}_{\tau, \boldsymbol{\epsilon}, \mathbf{z}} \big[ \| \mathcal{G}(\mathbf{z}^\tau, \tau, \mathbf{c}) - \mathbf{z} \|_2^2 \big],
\end{equation}
where $\tau \sim \mathcal{U}[0,1]$.

For real-time streaming, we enforce strict causality via causal attention masking.
At inference, we generate motion autoregressively by maintaining a history buffer of previously predicted latents.
Rather than conditioning on past motion explicitly—which led to mode collapse—we enforce temporal consistency through imputation.
Given the predicted history $\mathbf{z}_{1:k-1}$, we compute the corresponding noisy latents via Eq.~\ref{eq:z_tau} and sample fresh noise for the remaining sequence.
At each denoising step, we replace the noisy history tokens with their imputed values before proceeding.
After denoising, we append the newly predicted latent to the history buffer and slide forward by one block.

\subsection{Controllable Gaze Guidance}
Eye contact is a key non-verbal cue: more signals engagement, while less may indicate reserve.
However, appropriate eye contact varies widely---depending on preference, social context, and cultural norms.
This variability motivates making gaze behavior explicitly controllable at inference time.
While conditioning on user position enables plausible reactive motion, it restricts output to the gaze distribution in training data (Sec.~\ref{sec:datasets}).
To provide finer control, we introduce a tunable gaze guidance mechanism that modulates eye contact intensity based on user preference.

\begin{figure}[t]
\centering
\includegraphics[width=\columnwidth]{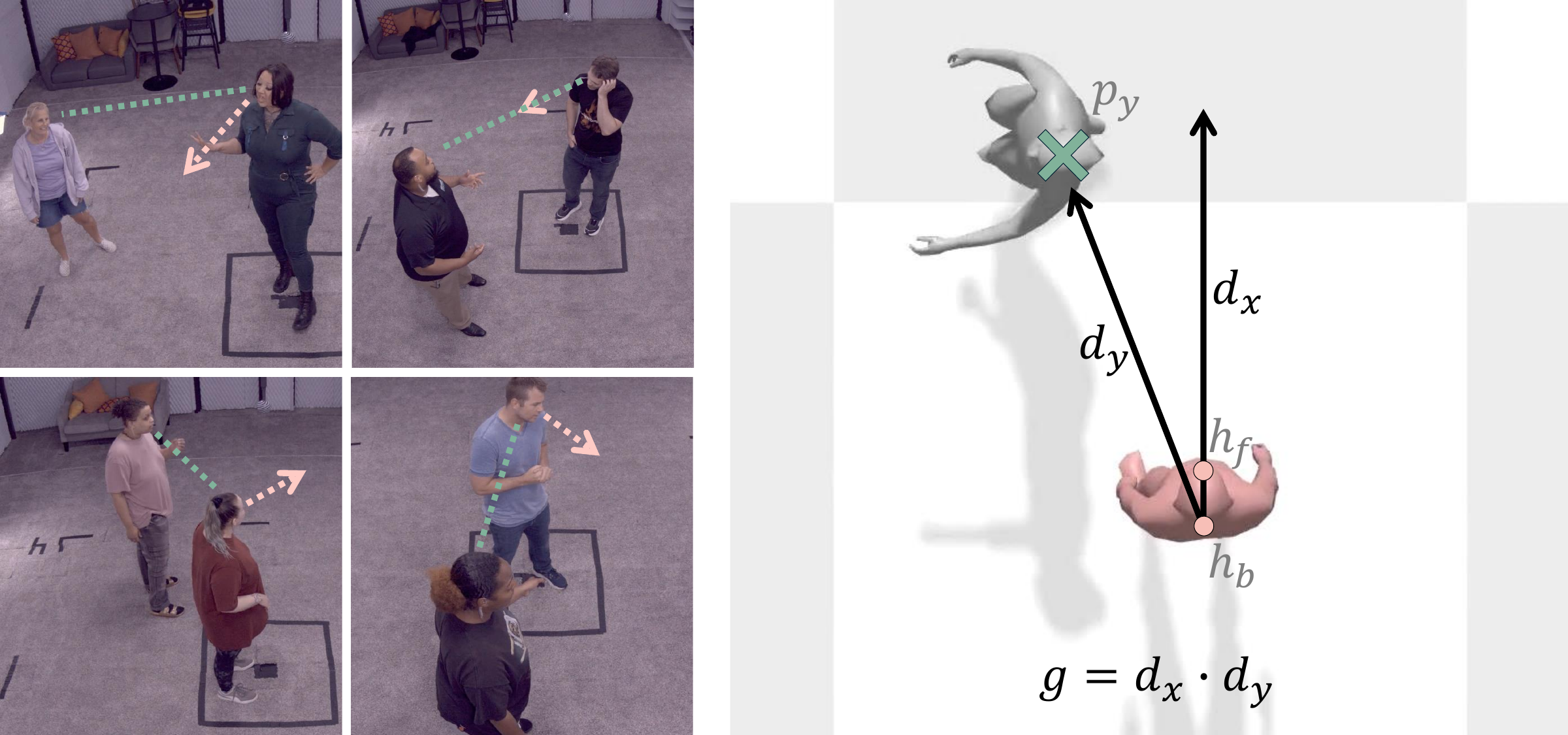}
\caption{Our training data spans a wide range of gaze behaviors, from sustained eye contact to complete gaze aversion (left). 
To enable controllable gaze at inference, we compute a gaze score $g$, where $\mathbf{d}_x$ is the agent's facing direction and $\mathbf{d}_y$ points toward the user (right). 
The score approaches $1$ when facing the user directly and $-1$ when facing away.}
\label{fig:gaze}
\end{figure}

We encode gaze based on head orientation relative to user position (Fig.~\ref{fig:gaze}).
Let $h_f, h_b \in \mathbb{R}^3$ denote the front and back of the agent's head.
We define the agent's facing direction as:
\begin{equation}
d_x = \frac{h_f - h_b}{\|h_f - h_b\|},
\end{equation}
and the direction toward the user as:
\begin{equation}
d_y = \frac{p_y - h_b}{\|p_y - h_b\|},
\end{equation}
The gaze score is then the dot product between these unit vectors:
\begin{equation}
g = d_x \cdot d_y.
\end{equation}
Intuitively, $g$ approaches $1$ when the agent faces the user directly, $0$ when looking perpendicular, and $-1$ when facing away.
Maximizing eye contact corresponds to maximizing $g$. 

During training, we concatenate the per-frame gaze score $\mathbf{g} \in \mathbb{R}^{T \times 1}$ with the conditioning $\mathbf{c} = [\mathbf{p}_y; \mathbf{a}; \mathbf{b}; \mathbf{g}]$ along the channel dimension, and apply classifier-free guidance by dropping $\mathbf{g}$ with 5 percent probability.
At inference, we specify a target gaze score to control eye contact intensity.
Crucially, guidance gently steers output toward the desired gaze range while preserving natural aversions and variation, yielding realistic and diverse motion.

\subsection{Dyadic conversational dataset}
\label{sec:datasets}
We use the dyadic conversation subset of the Embody 3D dataset \cite{mclean2025embody3d}. 
This subset contains around 50 hours captured in a multiview dome. 
The conversations cover a vast array of topics, including casual conversations, work discussions, and social interactions.
The demographics are diverse across age groups, genders, and ethnicities.
We use the audio and 3D motion annotations from the dataset.

This is the first dataset to capture 3D spatial proxemics in conversation.
Prior monadic datasets such as Speech2Gesture~\cite{ginosar2019learning} and BEAT~\cite{liu2022beat} offer diverse motion but lack spatial context, capturing a single speaker in isolation.
Existing dyadic datasets such as Audio2Photoreal~\cite{ng2024audio} and Panoptic Studio~\cite{joo2019towards} capture two-person interactions, but participants remain stationary and always face one another.
In contrast, Embody 3D contains scenarios where individuals walk freely, shift positions, and engage in natural, dynamic conversations.
\begin{table*}
    \centering
    \setlength{\abovecaptionskip}{4pt}
    \setlength{\belowcaptionskip}{-8pt}
    \caption{\textbf{Comparison with baselines and ablations (abl.) on 2048 test sequences.} C = causal, R = real-time. S = speaking (544 seq.), NS = non-speaking (1504 seq.). $\uparrow$ higher is better, $\downarrow$ lower is better. $^\dagger$Reducible to 600 fps without quality degradation.}
    \label{tab:baselines}
    \ra{1.15}
    \resizebox{\textwidth}{!}{%
    \begin{tabular}{@{}c l cc | r rl rl rl rl rl @{}}
    \toprule
    & & & & \textbf{FPS} & \multicolumn{2}{c}{\textbf{FGD} $\downarrow$ {\scriptsize(m $\times 10^1$)}} & \multicolumn{2}{c}{\textbf{FGD}$_\textbf{acc}$ $\downarrow$ {\scriptsize($\times 10^5$)}} & \multicolumn{2}{c}{\textbf{Foot Slide} $\downarrow$ {\scriptsize$\in [0,1]$}} & \multicolumn{2}{c}{\textbf{Wrist Var} $\uparrow$} & \multicolumn{2}{c}{\textbf{Head Ang.} $\uparrow$ {\scriptsize$\in [-1,1]$}} \\
    \cmidrule(lr){6-7} \cmidrule(lr){8-9} \cmidrule(lr){10-11} \cmidrule(lr){12-13} \cmidrule(lr){14-15}
    & & C & R & & Avg & {\scriptsize\textcolor{gray}{NS/S}} & Avg & {\scriptsize\textcolor{gray}{NS/S}} & Avg & {\scriptsize\textcolor{gray}{NS/S}} & Avg & {\scriptsize\textcolor{gray}{NS/S}} & Avg & {\scriptsize\textcolor{gray}{NS/S}} \\
    \midrule
    & \textit{GT} & -- & -- & -- & -- & {\scriptsize\textcolor{gray}{--}} & -- & {\scriptsize\textcolor{gray}{--}} & 0.01 & {\scriptsize\textcolor{gray}{0.01/0.01}} & 137.6 & {\scriptsize\textcolor{gray}{122.3/179.7}} & 0.81 & {\scriptsize\textcolor{gray}{0.80/0.84}} \\
    & & & & & & & & & & & & & & \\[-8pt]
    \multirow{5}{*}{\rotatebox{90}{\textit{Baselines}}} 
    & \cellcolor{gray!8}Random & \cellcolor{gray!8}\xmark & \cellcolor{gray!8}\cmark & \cellcolor{gray!8}4K & \cellcolor{gray!8}1.06 & \cellcolor{gray!8}{\scriptsize\textcolor{gray}{0.30/0.28}} & \cellcolor{gray!8}1.83 & \cellcolor{gray!8}{\scriptsize\textcolor{gray}{0.31/3.38}} & \cellcolor{gray!8} 0.01 & \cellcolor{gray!8}{\scriptsize\textcolor{gray}{0.01/0.01}} & \cellcolor{gray!8}188.1 & \cellcolor{gray!8}{\scriptsize\textcolor{gray}{190.5/181.6}} & \cellcolor{gray!8}0.28 & \cellcolor{gray!8}{\scriptsize\textcolor{gray}{0.27/0.32}} \\
    & \cellcolor{gray!8}NN & \cellcolor{gray!8}\xmark & \cellcolor{gray!8}\cmark & \cellcolor{gray!8}1K & \cellcolor{gray!8} 0.90 & \cellcolor{gray!8}{\scriptsize\textcolor{gray}{0.19/0.16}} & \cellcolor{gray!8} 0.77 & \cellcolor{gray!8}{\scriptsize\textcolor{gray}{0.02/0.51}} & \cellcolor{gray!8}0.01 & \cellcolor{gray!8}{\scriptsize\textcolor{gray}{0.01/0.01}} & \cellcolor{gray!8}97.0 & \cellcolor{gray!8}{\scriptsize\textcolor{gray}{85.7/128.2}} & \cellcolor{gray!8}0.59 & \cellcolor{gray!8}{\scriptsize\textcolor{gray}{0.57/0.64}} \\
    & \cellcolor{gray!8}MDM~\cite{tevet2022human} & \cellcolor{gray!8}\xmark & \cellcolor{gray!8}\xmark & \cellcolor{gray!8} 90 & \cellcolor{gray!8}3.48 & \cellcolor{gray!8}{\scriptsize\textcolor{gray}{1.93/2.66}} & \cellcolor{gray!8}2.88 & \cellcolor{gray!8}{\scriptsize\textcolor{gray}{0.64/5.37}} & \cellcolor{gray!8}0.11 & \cellcolor{gray!8}{\scriptsize\textcolor{gray}{0.11/0.11}} & \cellcolor{gray!8}61.4 & \cellcolor{gray!8}{\scriptsize\textcolor{gray}{57.9/71.0}} & \cellcolor{gray!8}0.81 & \cellcolor{gray!8}{\scriptsize\textcolor{gray}{0.80/0.84}} \\
    & \cellcolor{gray!8}A2P~\cite{ng2024audio} & \cellcolor{gray!8}\xmark & \cellcolor{gray!8}\xmark & \cellcolor{gray!8} 90 & \cellcolor{gray!8} 2.01 & \cellcolor{gray!8}{\scriptsize\textcolor{gray}{0.54/0.80}} & \cellcolor{gray!8} 2.31 & \cellcolor{gray!8}{\scriptsize\textcolor{gray}{0.43/4.95}} & \cellcolor{gray!8} 0.02 & \cellcolor{gray!8}{\scriptsize\textcolor{gray}{0.02/0.02}} & \cellcolor{gray!8} 69.4  & \cellcolor{gray!8}{\scriptsize\textcolor{gray}{59.0/98.2}} & \cellcolor{gray!8}0.71 & \cellcolor{gray!8}{\scriptsize\textcolor{gray}{0.70/0.73}} \\
    & \cellcolor{gray!8}SHOW~\cite{yi2023generating} & \cellcolor{gray!8}\cmark & \cellcolor{gray!8}\cmark & \cellcolor{gray!8} 230 & \cellcolor{gray!8}1.99 & \cellcolor{gray!8}{\scriptsize\textcolor{gray}{0.65/0.77}} & \cellcolor{gray!8}2.22 & \cellcolor{gray!8}{\scriptsize\textcolor{gray}{0.02/8.10}} & \cellcolor{gray!8}0.27 & \cellcolor{gray!8}{\scriptsize\textcolor{gray}{0.26/0.32}} & \cellcolor{gray!8}65.0 & \cellcolor{gray!8}{\scriptsize\textcolor{gray}{58.0/84.4}} & \cellcolor{gray!8}0.61 & \cellcolor{gray!8}{\scriptsize\textcolor{gray}{0.60/0.64}} \\
    & & & & & & & & & & & & & & \\[-8pt]
    \multirow{2}{*}{\rotatebox{90}{\textit{Abl.}}} & Ours in Joint Space (IK) & \cmark & \cmark & 300 & 2.35 & {\scriptsize\textcolor{gray}{0.40/0.81}} & 2.26 & {\scriptsize\textcolor{gray}{0.01/7.93}} & 0.03 & {\scriptsize\textcolor{gray}{0.03/0.04}} & 87.1 & {\scriptsize\textcolor{gray}{80.4/105.7}} & 0.72 & {\scriptsize\textcolor{gray}{0.71/0.75}} \\
    & Ours w/o VAE & \cmark & \cmark & 150 & 1.95 & {\scriptsize\textcolor{gray}{0.42/0.76}} & 2.24 & {\scriptsize\textcolor{gray}{0.01/8.08}} & 0.01 & {\scriptsize\textcolor{gray}{0.01/0.01}} & 96.9 & {\scriptsize\textcolor{gray}{90.3/115.2}} & 0.78 & {\scriptsize\textcolor{gray}{0.77/0.81}} \\
    & & & & & & & & & & & & & & \\[-8pt]
    & \cellcolor{gray!8}\textbf{Ours} & \cellcolor{gray!8}\cmark & \cellcolor{gray!8}\cmark & \cellcolor{gray!8} $300^\dagger$ & \cellcolor{gray!8}1.28 & {\scriptsize\textcolor{gray}{\cellcolor{gray!8}{0.35/0.87}}} & \cellcolor{gray!8}2.19 & {\scriptsize\textcolor{gray}{\cellcolor{gray!8}{0.01/7.81}}} & \cellcolor{gray!8}0.01 & \cellcolor{gray!8}{\scriptsize\textcolor{gray}{0.01/0.01}} & \cellcolor{gray!8}105.0 & \cellcolor{gray!8}{\scriptsize\textcolor{gray}{90.1/146.2}} & \cellcolor{gray!8}0.83 & \cellcolor{gray!8}{\scriptsize\textcolor{gray}{0.82/0.85}} \\
    \bottomrule
    \end{tabular}%
    }
    \end{table*}

    \begin{table}
    \centering
    \setlength{\abovecaptionskip}{4pt}
    \setlength{\belowcaptionskip}{-8pt}
    \caption{\textbf{Effect of gaze control on motion. $\emptyset$ denotes that gaze control is disabled.}}
    \label{tab:gaze}
    \ra{1.15}
    \resizebox{\columnwidth}{!}{%
    \begin{tabular}{@{}c rrrrr@{}}
    \toprule
    $g$ & \textbf{FGD} $\downarrow$ & \textbf{FGD}$_\textbf{acc}$ $\downarrow$ & \textbf{Foot Slide} $\downarrow$ & \textbf{Wrist Var} $\uparrow$ & \textbf{Head Ang.} $\uparrow$ \\
    \midrule
    \rowcolor{gray!8}
    $\emptyset$ & 1.28 & 2.19 & 0.01 & 105.0 & 0.83 \\
    0.0 & 0.99 & 2.18 & 0.01 & 111.1 & 0.56 \\
    \rowcolor{gray!8}
    0.8 & 0.92 & 2.19 & 0.01 & 110.8 & 0.76 \\
    1.0 & 1.49 & 2.20 & 0.01 & 106.6 & 0.96 \\
    \rowcolor{gray!8}
    \bottomrule
    \end{tabular}%
    }
    \end{table}


\section{Experiments}
\label{sec:experiments}

We evaluate our model's ability to generate realistic, spatially-aware conversational motion. 
Following prior works~\cite{ng2024audio,yi2023generating}, we quantitatively measure realism and diversity against ground truth, and additionally assess gaze alignment to determine whether the agent appropriately orients toward the user within the distribution of natural conversational behavior.
Our results show that our model generates motion competitive with state-of-the-art methods—including non-causal, non-real-time approaches—while being both causal and real-time. \textbf{For qualitative results, please refer to the Supp.~Video}.

\paragraph{Implementation Details}
We train our model and run all experiments on an A100 GPU. 
For all experiments, we set the sequence length $T=400$. 
Videos are sampled at 30 fps while the audio is sampled at 48kHz. 
For the motion representation, we use MHR~\cite{MHR:2025} which allows us to render photorealistic avatars.
For our VAE, we a stride of $s=4$, and the encoder and decoder each have 9 layers with 4 attention heads and a hidden dimension of 256. We set $\beta=1e^{-4}$ for the KL loss.
For the flow matching model, we encode each modality using a learned positional encoding before concatenating them along the channel dimension. We then use rope for temporal positional encoding. 
To incorporate the noise timestep, we use AdaLNZero. 
We use 4 transformer layers with 4 attention heads and hidden dimension of 1024. 
We train with local batch size of 16 across 8 gpu's. 
During inference, we use a cfg of 1.3 to control the conditioning strength.
Since not all methods are autoregressive or causal, we calculate the method's fps by generating all 400 frames in one go and then dividing the total time taken by 400.

\paragraph{Evaluation Metrics}
We evaluate motion along five axes:
\begin{enumerate}
    \item \textbf{FGD} (Fréchet Gesture Distance), which measures distributional similarity between generated and ground-truth poses via the Fréchet distance over the vertex positions of the mesh;
    \item \textbf{FGD}$_\textbf{acc}$, the same metric computed on acceleration to assess motion smoothness and dynamics;
    \item \textbf{Foot Slide}, the fraction of frames where feet are near the ground ($<$5\,cm) yet moving horizontally ($>$3\,cm/s), indicating skating artifacts;
    \item \textbf{Wrist Var}, the average wrist velocity measuring gesture expressiveness; and
    \item \textbf{Head Ang.}, the mean dot product between the agent's facing direction and the vector toward the user, quantifying gaze alignment ($1$\,=\,facing user, $-1$\,=\,facing away).
\end{enumerate}
We classify each clip as speaking (S) or non-speaking (NS) based on the agent's audio energy, and report both an overall average and separate S/NS values for each metric to enable analysis across conversational contexts.
For most metrics, the average reflects a weighted mean of the S and NS values.
However, for FGD and FGD$_\text{acc}$, the computation differs: the Avg column reports the mean of per-batch Fréchet distances, whereas the S and NS values are each computed by first pooling all clips of that category across all batches, then measuring a single Fréchet distance on the pooled distribution.
This pooling is necessary because individual batches may contain too few clips of one category for reliable covariance estimation.
As a consequence, the per-batch averages systematically exceed the pooled S/NS values due to small-sample-size bias in covariance estimation, and the Avg is not a simple weighted combination of S and NS.
Note that FGD$_\text{acc}$ is substantially higher for speaking clips than non-speaking clips, reflecting increased gestural dynamics during speech.

\paragraph{Baselines and Ablations}
Since no prior work addresses real-time, spatially-aware conversational motion generation, we cannot directly compare against existing methods.
To ensure a fair comparison, we retrain all prior works on our dataset and motion representation (Sec.~\ref{sec:motion-representation}).
We deliberately select foundational architectures—diffusion-based, VQ-based, and hybrid methods—that underpin many recent state-of-the-art systems, rather than task-specific variants with additional modules (e.g., text encoders or domain-specific losses).
This ensures a fair comparison of core generative capabilities.
We compare against:
\begin{itemize}
    \item \textbf{Random}: Randomly samples a motion sequence from the training set, providing a lower bound on performance.
    \item \textbf{NN}: A nearest-neighbor retrieval baseline that selects motion based on the conditioning inputs. 
        For audio matching, we use HuBERT embeddings. 
        We use a library of 2048 motion sequences randomly sampled from the training set and match across the full clip rather than via sliding windows, which yielded better temporal coherence and overall performance.
    \item \textbf{MDM}~\cite{tevet2022human}: A diffusion-based model originally designed for text-conditioned motion generation that has since become a foundation for many subsequent methods that have extended it to support various conditioning signals.
        We adapt MDM to use the same conditioning inputs for our domain: agent audio, user audio, and user head trajectory.
        It operates non-causally and does not run in real-time.
    \item \textbf{A2P}~\cite{ng2024audio}: A hybrid approach combining VQ-based discrete representations with diffusion-based refinement. 
        It operates autoregressively but is not real-time due to its multi-stage pipeline.
    \item \textbf{SHOW}~\cite{yi2023generating}: A VQ-based autoregressive model designed to generate upper-body, conversational 3D motion from speech.
        It employs separate VQ-VAEs for arm and hand movements, followed by an autoregressive generator for full upper-body motion.
        With minimal modification to the original architecture, we condition SHOW on agent audio alone to evaluate how existing audio-only methods perform in spatially-aware settings.
\end{itemize}

We also run ablation studies to isolate two key design choices: our motion representation and latent compression via the VAE.
\begin{itemize}
    \item \textbf{Ours in Joint Space (IK)}: Instead of our Euclidean representation (Sec.~\ref{sec:motion-representation}), we encode traditional joint angles with the VAE. Mesh positions are then recovered via inverse kinematics.
    \item \textbf{Ours w/o VAE}: We remove the causal VAE, directly predicting Euclidean positions from the transformer.
\end{itemize}

\subsection{Quantitative Results}
Tab.~\ref{tab:baselines} summarizes our main results across five evaluation axes.
We organize our analysis by first examining retrieval baselines, then generative baselines, and finally our ablations.

\paragraph{Retrieval Baselines (Random, NN)}
The retrieval baselines achieve the lowest FGD scores (\textbf{Random}: 1.06, \textbf{NN}: 0.90) since they sample directly from the true data distribution—outperforming \textbf{Ours} (1.28) on this metric alone.
However, this advantage is superficial: retrieval methods cannot jointly satisfy all criteria.
\textbf{Random}'s gaze alignment score (0.28) is catastrophic compared to \textbf{Ours} (0.83) since randomly sampled motion bears no relation to user position.
\textbf{NN} addresses this by jointly matching audio features (HuBERT embeddings) and user position, improving the gaze alignment to 0.59.
While better than \textbf{Random}, this still falls short of \textbf{Ours} (0.83) for two reasons: (1) jointly matching audio and spatial features is non-trivial, as optimizing for one may compromise the other, and (2) no clip in the dataset exactly matches the target user trajectory.
While both retrieval methods achieve near-zero foot sliding (0.01) by copying real motion (matching \textbf{Ours}), their wrist variance reveals further limitations: \textbf{Random} (188.1) overshoots GT (137.6) due to context-agnostic sampling, while \textbf{NN} (97.0) undershoots as retrieval favors common, less expressive clips.
\textbf{Ours} (105.0) strikes a better balance.
These results highlight a key distinction: while retrieval achieves strong distributional metrics by construction, it is fundamentally limited to what exists in the dataset.
\textbf{Ours} instead generates novel motion that jointly optimizes for all criteria—achieving competitive FGD (1.28) while dramatically improving spatial awareness (0.83 vs.~\textbf{NN}'s 0.59).

\paragraph{Generative baselines}
To evaluate against non-real-time state-of-the-art in the dyadic (two-person) setting, we adapt \textbf{MDM} and \textbf{A2P} to use the same user-aware conditioning as \textbf{Ours}: agent audio, user audio, and user head trajectory.
When naively adapted to our domain, \textbf{MDM} achieves the worst FGD (3.48) among all methods.
Analysis reveals that \textbf{MDM} produces over-smoothed motion: its wrist variance (61.4) is only 45\% of \textbf{GT} (137.6), indicating severely dampened gestures.
This likely reflects an architecture mismatch: \textbf{MDM} was designed for text-to-motion with coarse action descriptions, not fine-grained audio-gesture synchronization and may favor global motion coherence over local dynamics.
\textbf{MDM} appropriately matches the ground truth gaze alignment (0.81) perfectly—perhaps due to its non-causal architecture having access to future user positions to allow it to preemptively react accordingly.
In contrast, \textbf{Ours} achieves similar gaze alignment (0.83) while operating causally, demonstrating that gaze alignment can be learned without requiring future information.
\textbf{MDM} also exhibits significant foot sliding (0.11), suggesting that diffusion directly over the euclidean representation actually struggles to maintain physical constraints without a learned latent prior.

\textbf{A2P} extends \textbf{MDM} with an additional VQ-based stage: discrete tokens are first generated autoregressively, then refined via diffusion.
This two-stage approach reduces FGD and foot sliding compared to \textbf{MDM}.
However, \textbf{A2P} still falls short of \textbf{Ours} across all metrics: higher FGD (2.01 vs.~1.28), lower wrist variance (69.4 vs.~105.0), and weaker gaze alignment (0.71 vs.~0.83).
Qualitatively, we find that A2P's coarse VQ keyframes can lag temporally, forcing the diffusion stage to correct for misaligned targets. 
This results in dampened gestures (lower wrist variance) and temporally offset gaze (lower gaze alignment).
Both diffusion methods also run at only 90 FPS—3$\times$ slower than \textbf{Ours}—and their reliance on future context prevents deployment in streaming applications.

\begin{figure}[t]
\centering
\includegraphics[width=\columnwidth]{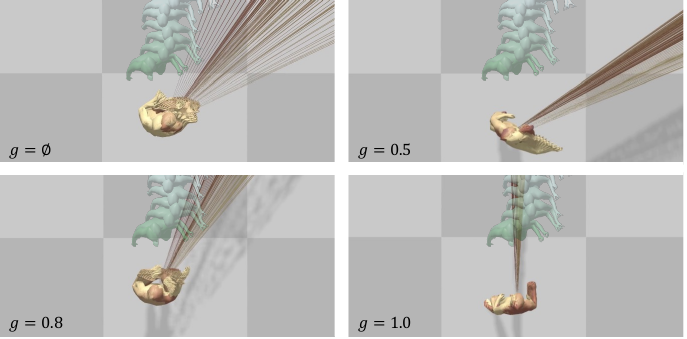}
\caption{We visualize the agent’s facing direction via projected lines (agent: yellow $\rightarrow$ red; user: blue $\rightarrow$ green). With no alignment $g=\emptyset$, the agent’s gaze is more diverse; as we increase $g$, the agent increasingly turns towards the user.}
\label{fig:gaze-eval}
\end{figure}

Unlike the diffusion methods, \textbf{SHOW} operates causally at 230 FPS, making it the most architecturally comparable baseline to \textbf{Ours}.
We evaluate it without user conditioning to serve as a monadic (single-agent) baseline.
However, \textbf{SHOW} struggles even in its original domain—suggesting fundamental architectural limitations even without user conditioning.
On foot sliding, the gap is stark: \textbf{SHOW} (0.27) is 27$\times$ worse than \textbf{Ours} (0.01), likely due to its separate VQ-VAEs for arms and hands—originally designed for upper-body motion—which lack body-ground coordination when extended to full-body generation.
On expressiveness, \textbf{SHOW}'s wrist variance (65.0) falls well below \textbf{Ours} (105.0).
Qualitatively, \textbf{SHOW} produces sweeping gestures but struggles with the rapid, fine-grained motion important for expressive speech—dynamics that \textbf{Ours} captures through its flow-based formulation.
As expected, the largest gap is in spatial awareness: \textbf{SHOW}'s gaze alignment (0.61) falls well below \textbf{Ours} (0.83).
This highlights a key limitation of audio-only conditioning—the audio signal does not encode user position, so the model cannot learn appropriate orientation.
\textbf{Ours} addresses this directly through explicit user conditioning, enabling spatially-aware generation.

\paragraph{Ablations}
We isolate the contributions of key design choices.
\textbf{Ours in Joint Space (IK)} replaces our Euclidean surface-point representation with traditional joint angles, requiring inverse kinematics to recover mesh positions.
A core issue is that joint-angle predictions face inherent ambiguity—multiple configurations can produce similar end-effector positions.
This directly impacts metrics that depend on precise positioning: gaze alignment drops from 0.83 to 0.72 (head orientation), and foot sliding increases from 0.01 to 0.03 (foot-ground contact).
The ambiguity may also encourage conservative predictions, which is reflected in wrist variance decreasing from 105.0 to 87.1—the model produces less expressive motion when end-effector targets are uncertain.
These results motivate our Euclidean surface-point approach, which directly specifies end-effector positions without ambiguity.

\textbf{Ours w/o VAE} removes the causal VAE, directly predicting motion from the transformer.
Without the VAE's learned latent structure, the model must predict high-dimensional motion directly, making it harder to capture the true motion distribution—FGD rises from 1.28 to 1.95.
However, physical plausibility metrics remain stable: foot sliding stays at 0.01 and wrist variance (96.9) remains comparable to \textbf{Ours} (105.0).
This indicates that the VAE's primary benefit is distributional—matching the motion manifold—rather than enforcing physical constraints, which our Euclidean representation seems to handle.
Inference speed also halves (300 to 150 FPS), as predicting in the compressed latent space is more efficient than directly generating high-dimensional motion.

\subsection{Gaze Control}
We evaluate gaze controllability by varying the guidance parameter $g$ at test time and applying classifier-free guidance to enforce the desired alignment (Tab.~\ref{tab:gaze}).
As shown in Fig.~\ref{fig:gaze-eval}, increasing $g$ from 0.0 (looking away) to 1.0 (always facing the user) also increases gaze alignment accordingly (0.56 $\rightarrow$ 0.96). This confirms our method's ability to explicitly control over agent orientation.
At $g{=}0.8$, which best matches ground truth (0.81), we even outperform the default no-guidance case ($\emptyset$) with lower FGD (0.92 vs.\ 1.28) and slightly higher wrist variance.
This suggests that moderate gaze guidance provides useful spatial grounding that improves overall motion quality.
At $g{=}1.0$, gaze alignment reaches 0.96 but FGD rises to 1.49, reflecting the trade-off between strict gaze adherence and natural motion variation.
At $g{=}0.0$, gaze alignment drops to 0.56 rather than zero, since complete aversion is rare in the dataset—the agent turns considerably away from the user but still adheres to the learned distribution.

\section{Conclusion}
\label{sec:conclusion}

We presented the first method for spatially-aware conversational motion, enabling virtual agents to orient toward and react to a moving user \emph{in real-time} while producing natural, speech-aligned gestures.
The architecture pairs a novel causal transformer-based VAE with a flow matching model conditioned on user trajectory and dyadic audio.
Recognizing that gaze preferences vary, we introduce a gaze alignment score steered via classifier-free guidance, decoupling learning from control.
Experiments show state-of-the-art quality at over 300 FPS, outperforming non-causal baselines 3$\times$ slower.
The causal, real-time nature enables deployment in streaming headset environments.

Our method inherits training data biases: underrepresented spatial configurations or gaze behaviors may generalize poorly.
While we demonstrate controllable gaze, other behaviors—gesture style, locomotion—are not yet controllable.
Extending to multi-party conversations would require architectural modifications.

\begin{acks}
  We would like to thank the Embody 3D team for making this project possible. We would also like to thank Abhay Mittal, Anastasis Stathopoulos, and Ethan Weber for helpful discussions. Thank you, Vasu Agrawal, Martin Gleize, and Srivathsan Govindarajan for making the demo possible.
\end{acks}

\clearpage
\begin{figure*}[p]
  \centering
  \includegraphics[width=\textwidth,height=\textheight,keepaspectratio]{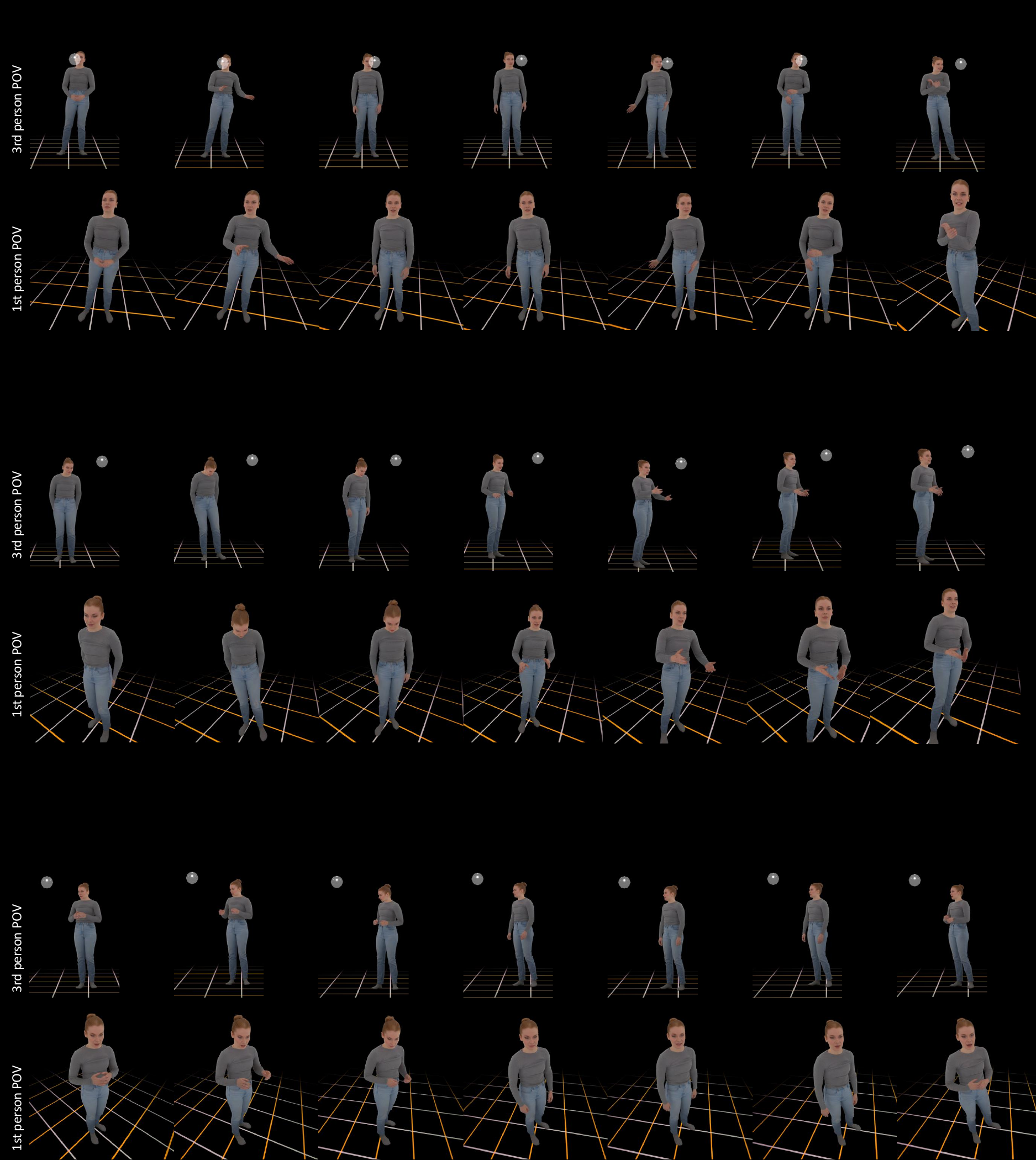}
  \caption{Sequences from our real-time demo system, rendered with a photorealistic avatar. The top row visualizes the user's headset location as a silver sphere. The bottom row shows the generated avatar from the user's (headset) viewpoint. Our method generates realistic conversational motion that is responsive to the user's spatial motion. Full videos are available on our \href{https://evonneng.github.io/sarah/}{project page}.}
  \label{fig:qualitative_fullpage}
\end{figure*}
\clearpage

\bibliographystyle{ACM-Reference-Format}
\bibliography{main}

\appendix

\section{Supplementary Material}
\label{sec:suppl}

\subsection{Video Results}
Please refer to \href{https://evonneng.github.io/sarah/arxiv_videos/full-sequence-d-web.mp4}{5 minute video} for this section. 

We start with the problem setup (00:00 -- 00:45) for a dyadic conversation between a user and an agent.
Given the user's 3D position and dyadic audio (from both user and agent), our goal is to generate spatially-aware 3D motion for the agent that aligns with the conversation and moves according to the user's 3D position.
From the generated motion, we can then render a photorealistic avatar.
\emph{Our model is lightweight and fast enough to enable streaming, allowing real-time interaction with the AI agent on VR platforms.}

The streamed results (00:46 -- 01:25) demonstrate that our model produces conversationally-appropriate gestures while naturally turning toward the user to signal social engagement.
The agent seamlessly transitions between speaking and listening modes, maintaining dyanamic gestures when speaking, and engaged idle gestures when listening.

Our method generalizes across diverse emotional contexts, producing contextually-appropriate body language: 
hands on hips and looking down when stressed or rejected (01:26 -- 02:00), 
lively gestures when excited (02:01 -- 02:26), 
clenched fists when angry (02:27 -- 02:41), 
and exaggerated bowing in celebratory agreement (02:41 -- 02:57).

To ensure that our model is controllable when it comes to gaze preferences, we also include a gaze score which we can tune at test time.
For lower gaze scores, the agent avoids direct facing the user. For the exact same input conditioning, increasing the gaze score results in more direct facing (02:58 -- 03:21).
When we fully drop out the gaze score ($g = 0$), the agent's gaze just follows whatever is in-distribution with the training dataset (03:22 -- 03:38).

We also compare against existing methods. Compared to MDM \cite{tevet2022human} our method produces considerably more lively gestures (03:39 -- 03:52).
Compared to Audio2Photoreal \cite{ng2024audio}, our method produces more realistic motion (03:53 -- 04:07). 
For Audio2Photoreal, it seems as if the VQ will predict slightly delayed motion which forces the diffusion side to catch up with, which results in distored motion.
Compared to TalkSHOW \cite{yi2023generating}, our method produces less motion artifacts since we predict the full-body motion in a single model (04:08 -- 04:22). 
Instead, TalkSHOW's VQ-based approach results in distored wrist motion artifacts and ample foot sliding.

The real-time nature of our model enables fully interactive AI agents in VR (04:23 -- end).
We generate dyadic conversations using off-the-shelf LLMs paired with text-to-speech models---here, ChatGPT for dialogue and Kyutai for speech synthesis.
This enables applications ranging from entertainment (e.g., gaming NPCs) to personal assistants.

\subsection{Training Details}
\label{sec:suppl-training}

We provide additional training hyperparameters and details not included in the main text.

\paragraph{Optimization}
We use the AdamW optimizer with $\beta_1 = 0.9$, $\beta_2 = 0.999$, and weight decay $1\times10^{-4}$.
The learning rate follows a linear warmup over the first 1,000 training steps, peaking at $1\times10^{-4}$.
The VAE is trained for 200K iterations before freezing, after which the flow matching model is trained for an additional 300K iterations.

\paragraph{Data Processing}
We use a 80/10/10 split for training/validation/test.
During training, we randomly sample a full sequence from the training set and from there, randomly sample a subsequence of length $T=400$ frames.
For test time, we use a sliding window of length $T=400$ and no overlap. We evaluate across the full set and generate 2048 sequences in total.

For audio features, we use HuBERT-Large, which is not fully causal. So at training time, we essentially do have some information leakage.
In order to ensure that it is fully causal at test time, we implement the streaming logic such that we never pass into HuBERT any future frames to avoid this leakage.
Instead, we always implement a sliding window logic where we pass in the current context and then the previous $T-s$ frames.
We find that shifting to this fully causal approach at test time does not degrade performance.

\paragraph{Latent Dimension}
The VAE latent dimension is $D_z = 256$.
With stride $s=4$ and sequence length $T=400$, this produces $K=100$ latent tokens per sequence.

\subsection{Inference Details}
\label{sec:suppl-inference}

\paragraph{Streaming Protocol}
For real-time deployment, we generate motion in chunks of $s=4$ frames. 
We then keep the last 2 tokens and then remove all the prior ones. 
In essence, we generate a total of 8 frames at a time.
As discussed in the main text (\label{sec:motion-gen}), we inpaint the history frames to maintain temporal consistency.
For each chunk, we run using midpoint solver with 4 iterations (8 nfe steps).
In this setting, we are able to achieve 60 fps at test time, which allows us to achieve real-time streaming performance.

\paragraph{Photorealistic Rendering}
We follow \cite{bagautdinov2021driving}, a learning based method, to render photorealistic avatars from the generated joint parameter motions.
The model takes as input one frame of facial expression, one frame of body pose, and a viewpoint direction.
We use an off the shelf method to generate facial expression parameters from speech audio.
The model then outputs a registered geometry and view dependent texture, which is used to synthesize images via rasterization.
For further details, please refer to \cite{bagautdinov2021driving}.

\end{document}